\documentclass[conference]{IEEEtran}
\IEEEoverridecommandlockouts
\usepackage[english]{babel}
\usepackage{ifpdf}
\usepackage{cite} 
\usepackage{url}
\usepackage{hyperref}
\usepackage{color}
\usepackage{pgf, tikz, pgfplots}
\usetikzlibrary{shapes, arrows, automata}
\usepackage{caption} 
\usepackage{subcaption} 
\usepackage{amsmath}
\usepackage{amsfonts, amssymb, amsthm}
\usepackage{mathrsfs}
\usepackage{upgreek}
\usepackage{algorithm,algpseudocode}
\usepackage{enumerate}
\usepackage{multirow}
\usepackage{rotating}
\usepackage{needspace}

\newcommand{\myparagraph}[1]{\needspace{1\baselineskip}\medskip\noindent {\bf #1.}}

\input{mysymbol.sty}

\newcommand{\QED}{\hfill\ensuremath{\blacksquare}}

\def\Tr{\mathsf{T}}

\title{Gated Graph Convolutional Recurrent\\Neural Networks}
\author{Luana~Ruiz, Fernando~Gama~
        and~Alejandro~Ribeiro
\thanks{Supported by NSF CCF 1717120, ARO W911NF1710438, ARL DCIST CRA W911NF-17-2-0181, ISTC-WAS and Intel DevCloud. The authors are with the Dept. of Electrical and Systems Eng., Univ. of Pennsylvania. Email: \{rubruiz,fgama,aribeiro\}@seas.upenn.edu.
}
}

\begin{document}
%
\maketitle
\begin{abstract}
Graph processes model a number of important problems such as identifying the epicenter of an earthquake or predicting weather. In this paper, we propose a Graph Convolutional Recurrent Neural Network (GCRNN) architecture specifically tailored to deal with these problems. GCRNNs  use convolutional filter banks to keep the number of trainable parameters independent of the size of the graph and of the time sequences considered. We also put forward Gated GCRNNs, a time-gated variation of GCRNNs akin to LSTMs. When compared with GNNs and another graph recurrent architecture in experiments using both synthetic and real-word data, GCRNNs significantly improve performance while using considerably less parameters.
\end{abstract}
%


\section{Introduction} \label{sec:intro}


The availability of ever-growing volumes of data --- often referred to as \textit{big data} --- has propelled the use of neural network architectures in both engineering and less traditional fields, such as medicine \cite{lipton2015learning} and business consulting \cite{phillips2015business}. But learning from large datasets comes with a challenge: it requires complex models with many parameters which, on the one hand, are time and memory-intensive and, on the other, increase the risk of overfitting. To get around these issues, a lot of effort has been put into designing architectures that exploit the underlying structure of data using an amenable number of parameters. The first example are Convolutional Neural Networks (CNNs) \cite{kuo17-recos}, which use banks of convolutional filters whose number of parameters is independent of the size of the input to extract shared features across grid-like signals (e.g. images). Then, there are Graph Convolutional Neural Networks (GNNs) \cite{kipf17-classifgcnn, gama18-gnnarchit, defferrard17-cnngraphs, ruiz19-median}, which achieve the same purpose on graph data using graph convolutional filters also known as linear shift-invariant graph filters (LSI-GFs) \cite{segarra17-linear}. A third example are Recurrent Neural Networks (RNNs) \cite{goodfellow16-deeplearn}, designed to process sequential data through the addition of a state or memory variable that stores past information.

The sequences processed by RNNs are usually temporal processes, but they are rarely one-dimensional, i.e., they do not vary only in time. In particular, we will be interested in sequences that are best represented by \emph{graph processes} \cite{gama18-glln}. Graph processes model a variety of important problems; some illustrative examples are weather prediction from data collected at weather station networks \cite{perraudin2017stationary} and identifying the epicenter of an earthquake from seismic waves \cite{grassi2017time}.  

To deal with these scenarios, we introduce a Graph Convolutional Recurrent Neural Network (GCRNN) architecture where the hidden state is a graph signal computed from the input and the previous state using banks of graph convolutional filters and, as such, stored individually at each node. In addition to being local, in GCRNNs the number of parameters to learn is independent of time because the graph filters that process the input and the state are the same at every time instant. GCRNNs can take in graph processes of any duration, which gives control over how frequently gradient updates occur. They can also learn many different representations: a signal (whether supported on a graph or not) or a sequence of signals; a class label or a sequence of labels. While other graph-based recurrent architectures have been proposed in \cite{li2017diffusion, zhang2018gaan, yu2017spatio}, they are limited to representing sequences of graph signals and, in general, problem specific (most commonly to traffic forecasting). A fourth graph recurrent formulation has been introduced in \cite{ioannidis2018recurrent}, but it uses recurrence as a way of re-introducing the input data at each layer to capture multiple types of diffusion, and as such does not operate on graph processes. In this architecture, the number of learnable parameters also depends on the number of recurrent layers.

Our GCRNN architecture is further extended to include time gating variables analogous to the input and forget gates of Long Short Term Memory units (LSTMs) \cite{goodfellow16-deeplearn}, which are also implemented using GCRNNs. The objective of gating is twofold: on the input side, to control the importance given to new information, and on the state side, how much of the stored past information the model should ``forget''.

GCRNNs' ability to learn both graph and time dependencies and the importance of the gating mechanism for long input sequences are demonstrated in experiments on synthetic and real-world data. GCRNNs are compared with basic GNNs and with the DCRNN, a gated graph recurrent architecture from the existing literature \cite{li2017diffusion}. Numerical results show that: (i) GCRNNs largely improve upon GNNs when processing sequential graph data, and (ii) our model uses considerable less parameters and achieves better performance than the DCRNN. 


\section{Graph Processes} \label{sec:graphProcesses}



Let $\ccalG = (\ccalV, \ccalE, \ccalW)$ be a graph where $\ccalV$ is a set of $N$ nodes, $\ccalE \subseteq \ccalV \times \ccalV$ is the set of edges and $\ccalW: \ccalE \to \reals$ is a function that assigns weights to each edge. The topology of the graph $\ccalG$ can be described by a matrix $\bbS \in \reals^{N \times N}$ that captures the sparsity pattern of its structure, i.e., $[\bbS]_{ij}=s_{ij}$ can be nonzero only if $(j,i) \in \ccalE$ or $j=i$. Typical choices for this matrix in the literature are the adjacency \cite{sandryhaila13-dspg}, the Laplacian \cite{shuman13-mag}, the random walk \cite{Heimowitz17-MarkovGSP} and their normalized counterparts \cite{defferrard17-cnngraphs, kipf17-classifgcnn}. We define the neighborhood of node $i$ as $\ccalN_{i} = \{j \in \ccalV: (j,i) \in \ccalE\}$, and use the expression \textit{local operations} to refer to operations that can be computed by successive interactions of a node $i$ with its neighborhood $\ccalN_i$.

We model data as \emph{graph signals}: a graph signal $\bbx: \ccalV \to \reals$ is such that each element $[\bbx]_{i} = x_{i}$ corresponds to the value of the graph signal at node $i \in \ccalV$. The most basic interaction between the graph signal $\bbx$ and the graph $\ccalG$ is given by the operation $\bbS \bbx$, where, for each $i=1,\ldots,N$, we have
\begin{equation} \label{eqn:graphShift}
    [\bbS \bbx]_{i} = \sum_{j=1}^{N} [\bbS]_{ij} [\bbx]_{j} = \sum_{j \in \ccalN_{i}} s_{ij} x_{j}\ .
\end{equation}
The operation described by \eqref{eqn:graphShift} is local, because its output can be computed by interacting only with $\ccalN_{i}$ due to the sparsity pattern of $\bbS$. The matrix $\bbS$ has the effect of shifting around the graph the information contained in its nodes and is henceforth called the \emph{graph shift operator} (GSO).

We are interested in graph signals that change with time over the same graph support, typically known as \emph{graph processes} \cite{gama18-glln}. A graph process $\{\bbx_{t}\}_{t \in \mbN_{0}}$ is a sequence of graph signals $\bbx_{t} \in \reals^{N}$, which are all defined over the same graph support, $\bbx_{t}: \ccalV \to \reals$. More often than not, there exists some dependency relationship between graph signals at different time instants. This (causal) dependency can be described, generically, by $\bbx_{t} = f(\bbx_{t-1},\bbx_{t-2},\ldots)$ for some function $f$ that in practice is usually unknown.

The graph process is typically accompanied by some target representation $\ccalY$ relevant to the task at hand, which gives rise to pairs $(\{\bbx_{t}\}, \ccalY)$. In regression problems, the target representation is usually another sequence $\{\bby_{t}\}$, while in classification problems it is a single element $\bby$ characterizing the sequence over some time interval. The general objective of learning over sequences is to obtain a meaningful estimate $\hat{\ccalY}$ of the target representation $\ccalY$ using $\{\bbx_{t}\}$. To do this, we propose a novel architecture that we call the \emph{Graph Convolutional Recurrent Neural Network} (GCRNN).

To enhance the descriptive capabilities of our data model, in what follows we consider sequences comprised of $F$ different features $\bbx_{t}^{f}$ for $f=1,\ldots,F$, where each $\bbx_{t}^{f} \in \reals^{N}$ is a graph signal. A more compact representation is given by the matrix $\bbX_{t} \in \reals^{N \times F}$, where each column $\bbx_{t}^{f} \in \reals^{N}$ is a graph signal ($f=1,\ldots,F$) and each row $\tbx_{t}^{i} \in \reals^{F}$ gathers the feature values collected at a single node ($i=1,\ldots,N$)
\begin{equation}
    \bbX_{t} 
        = \left[\bbx_{t}^{1},\ldots,\bbx_{t}^{F}\right] 
        = \begin{bmatrix}
            (\tbx_{t}^{1})^{\Tr} \\
            \vdots \\
            (\tbx_{t}^{N})^{\Tr}
        \end{bmatrix}\ .
\end{equation}
While we have defined a local operation on $\bbx_{t}^{f}$ as one that respects the sparsity of the graph [cf. \eqref{eqn:graphShift}], we note that any operation on $\tbx_{t}^{i}$ can be called local as well, since it involves values that are already available at that node.


\section{Graph Convolutional Recurrent\\Neural Networks} \label{sec:archit}



A recurrent neural network (RNN) approximates the temporal dependencies of a sequence $\{\bbx_{t}\}$ using a hidden Markov model, i.e. $\bbx_{t+1} \approx g(\bbx_{t},\bbh_{t})$ for some function $g$ and some \emph{hidden state sequence} $\{\bbh_{t}\}$. The hidden state sequence is
\begin{equation} \label{eqn:RNN}
    \bbh_{t} = \sigma \left( \bbA \bbx_{t}+ \bbB \bbh_{t-1} \right)\ ,
\end{equation}
where $\bbA$ and $\bbB$ are linear transforms and $\sigma$ is a nonlinear function, so as to endow the RNN \eqref{eqn:RNN} with higher descriptive power. The representation estimate $\hat{\ccalY}$ can then be obtained from these hidden states. For instance, in a regression problem, we would have $\hby_{t} = \rho(\bbC \bbh_{t})$ with $\rho$ a nonlinear function and $\bbC$ a linear transform, and in a classification problem, $\hby = \rho(\bbC \bbh_{T})$ for the state $\bbh_{T}$ computed after some interval $T$.

The parameters of the linear transforms $\bbA$, $\bbB$ and $\bbC$ can be learned by minimizing some loss function $\ccalL(\ccalY, \hat{\ccalY})$ over a training set $\ccalT = \{(\{\bbx_{t}\}, \ccalY)\}$. We note that the learned linear transforms are the same for all $t$, giving the RNN \eqref{eqn:RNN} enough flexibility to adapt to sequences of different length. Likewise, the number of parameters is independent of the length of the sequence. The hidden state $\bbh_{t}$ is expected to store all the past information that is relevant for estimating the target representation.

The knowledge that the sequence $\{\bbx_{t}\}$ is comprised of graph signals defined over the same graph $\ccalG$ with GSO $\bbS$ can be exploited by forcing the linear transforms $\bbA$ and $\bbB$ to account for this structure,
\begin{equation} \label{eqn:GRNN}
    \bbh_{t} = \sigma \big( \bbA(\bbS) \bbx_{t}+ \bbB(\bbS) \bbh_{t-1} \big)\ .
\end{equation}
We name model \eqref{eqn:GRNN} a \emph{Graph Recurrent Neural Network} (GRNN). A particularly compelling parametrization of the linear transforms $\bbA(\bbS)$ and $\bbB(\bbS)$ is given by banks of linear shift-invariant graph filters (LSI-GFs). LSI-GFs don the GRNN model \eqref{eqn:GRNN} with convolutional characteristics, since they are permutation-invariant local operations and make the number of learnable parameters independent of the size of the graph. More precisely, let the hidden state be described by $D$ features, where each feature $\bbh_{t}^{d} \in \reals^{N}$ is a graph signal. We define the \emph{Graph Convolutional Recurrent Neural Network} (GCRNN) as
\begin{equation} \label{eqn:GCRNN}
    \bbh_{t}^{d} = \sigma \left( \sum_{f=1}^{F} \bbA^{df}(\bbS) \bbx_{t}^{f} + \sum_{d'=1}^{D} \bbB^{dd'}(\bbS) \bbh_{t-1}^{d'} \right)\ ,
\end{equation}
where $\bbA^{df}(\bbS)$ and $\bbB^{dd'}(\bbS)$ are the LSI-GFs
\begin{equation} \label{eqn:LSIGFs}
    \bbA^{df}(\bbS) = \sum_{k=0}^{K-1} a_{k}^{df} \bbS^{k} \quad , \quad \bbB^{dd'}(\bbS) = \sum_{k=0}^{K-1} b_{k}^{dd'} \bbS^{k}\ ,
\end{equation}
for $d,d' = 1,\ldots,D$ and $f=1,\ldots,F$. The \emph{filter taps} $\{a_{k}^{df}\}_{k=0}^{K-1}$ and $\{b_{k}^{dd'}\}_{k=0}^{K-1}$ are the learnable parameters of the linear transform. Note that there are $DFK + D^{2}K$ such parameters and that their number is independent of the sequence length and of the size of the graph $N$. Another primary feature of GCRNNs is that the LSI-GFs \eqref{eqn:LSIGFs} are local operations, since they can be computed by $K-1$ successive interactions with the neighbors of each node. The capacity of GCRNNs can be further increased (while maintaining the convolutional characteristics) by using graph convolutional neural networks \cite{gama18-gnnarchit} with several layers in place of $\bbA(\bbS)$ and $\bbB(\bbS)$ in \eqref{eqn:GRNN}.

Describing the hidden states $\bbh_{t}^{d}$ as a collection of $D$ graph signals lets us again exploit the graph structure in the computation of the target representation $\hat{\ccalY}$. Let $\bbH_{t} \in \reals^{N \times D}$ be a matrix where the hidden states $\bbh_{t}^{d} \in \reals^{N}$ are its columns,
\begin{equation}
    \bbH_{t} 
    = \left[\bbh_{t}^{1},\ldots,\bbh_{t}^{D}\right] 
    = \begin{bmatrix}
        (\tbh_{t}^{1})^{\Tr} \\
        \vdots \\
        (\tbh_{t}^{N})^{\Tr}
    \end{bmatrix}\ ,
\end{equation}
and where the rows of $\bbH_{t}$ collect the $D$ features at node $i$,  $\tbh_{t}^{i} \in \reals^{D}$ for $i=1,\ldots,N$. The estimated representation is computed as $\hby_{t} = \rho(\bbC(\bbS) \bbH_{t})$ for the regression problem and $\hby = \rho(\bbC(\bbS) \bbH_{T})$ for the classification problem. Operation $\bbC(\bbS)$ can be replaced by a graph convolutional neural network to exploit locality, followed by a fully connected layer to adapt dimensions when mapping $\bbH_{t}$ to $\hby_{t}$ or $\hby$.

The regression problem where the target representation sequence $\{\bby_{t}\}$ is a sequence of graph signals $\bby_{t}^{g} \in \reals^{N}$, with $g=1,\ldots,G$ denoting different features, is of particular interest, as it allows for two possible local models to compute $\hby_{t}^{g}$. The first possibility is to apply a LSI-GF to $\bbh_{t}^{d}$,
\begin{equation}
    \hby_{t}^{g} = \rho \left( \sum_{d=1}^{D} \bbC^{gd}(\bbS) \bbh_{t}^{d} \right) \quad , \quad \bbC^{gd}(\bbS) = \sum_{k=0}^{K-1} c^{gd}_{k} \bbS^{k}\ ,
\end{equation}
which demands $K-1$ successive interactions with the neighbors of each node. This requires $DGK$ learnable parameters. Alternatively, the second possibility is to estimate the target features at each node by applying a linear transformation to its own features,
\begin{equation} \label{eqn:local_mlp}
    \widehat{\tby}_{t}^{i} = \bbC \tbh_{t}^{i}
\end{equation}
where $\bbC \in \reals^{G \times D}$, for each $i=1,\ldots,N$. This alternative entails no neighbor interactions since $\tbh_{t}^{i}$ is stored at node $i$, and requires only $DG$ parameters if the same linear transform $\bbC$ is learned for all nodes.


\section{Gating} \label{sec:gating}

%
\begin{figure*}[t]
	\centering
	\begin{subfigure}{.95\columnwidth}
		\centering
		\includegraphics[width=.85\columnwidth, height=.16\textheight, keepaspectratio]{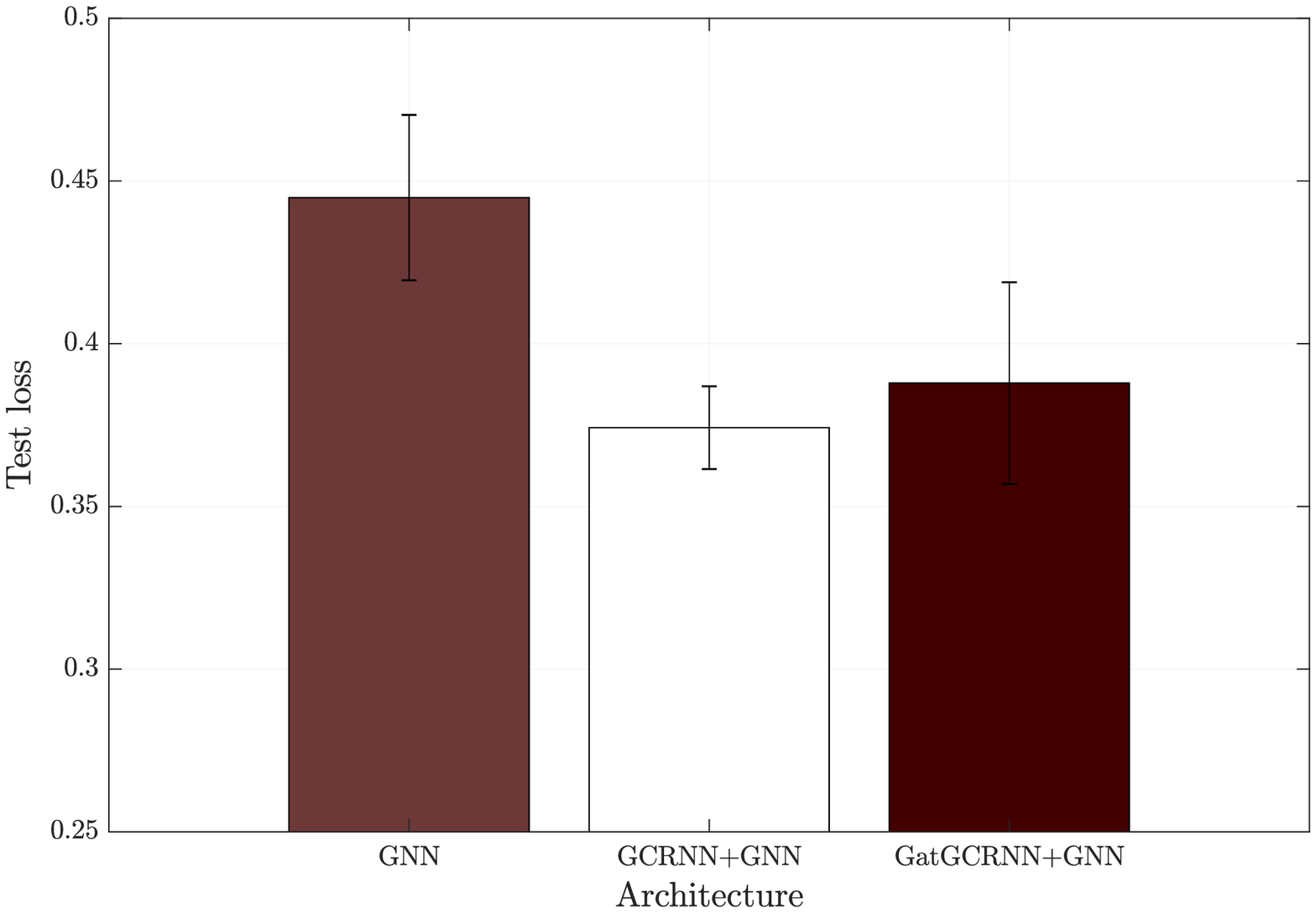}
		\caption{}
		\label{GNN}
	\end{subfigure}
	\hspace{0.25in}
	\begin{subfigure}{.95\columnwidth}
		\centering
		\includegraphics[width=.85\columnwidth, height=.16\textheight, keepaspectratio]{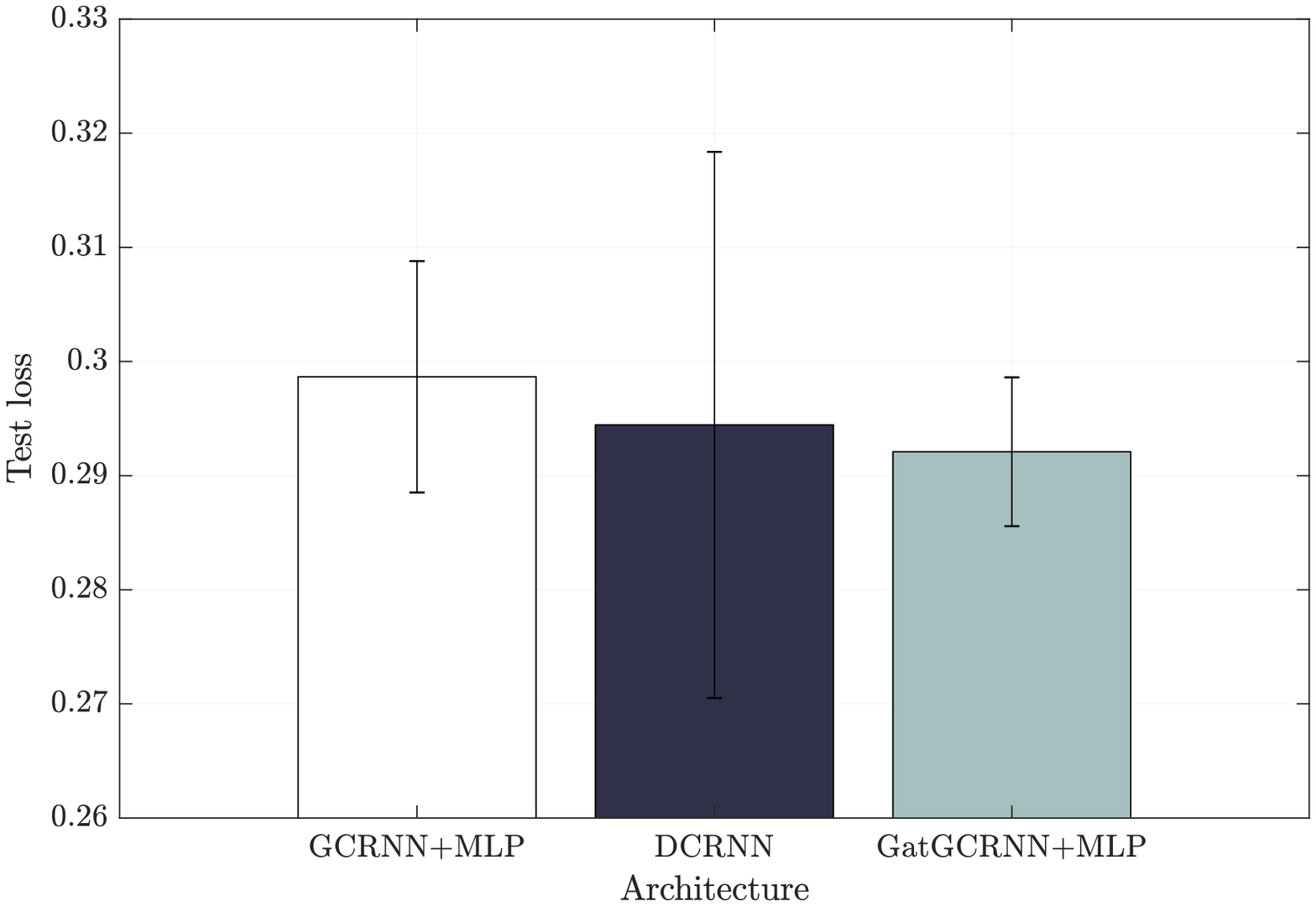}
		\caption{}
		\label{MLP}
	\end{subfigure}
	\caption{Test loss (mean absolute error) in 10-step prediction. Average loss and standard deviation on 10 different graphs and diffusion datasets. (\subref{GNN}) Test loss for a GNN with 2 convolutional layers (right), a GCRNN followed by a GNN (center) and a Gated GCRNN followed by a GNN (left). 
(\subref{MLP}) Test loss for a GCRNN followed by nodewise MLPs with shared parameters (right), the DCRNN from \cite{li2017diffusion} (center) and the gated GCRNN with localized MLPs (left). }
	\label{fig:synthetic}
\end{figure*}	
Deeper RNNs allow taking long term dependencies into account, but this often comes with the challenge of vanishing gradients as long term interactions get exponentially smaller weights at each training step \cite{goodfellow16-deeplearn}. This is addressed by time gating mechanisms such as the input and forget gates of Long Short-Term Memory units (LSTMs) and the reset and update gates of Gated Recurrent Units (GRUs). These gates are essentially variables taking values between 0 and 1 --- each one estimated by their own neural network model --- that multiply either or both the input and the state to control the amount of information passed through with time. 

Time gating can be readily extended to the context of graph processes. Based on the GCRNN model \eqref{eqn:GCRNN}, we define a time-gated architecture as follows,
\begin{equation} \label{eqn:GGCRNN}
    \bbh_{t}^d 
        = \sigma \bigg( \alpha_{t} \sum_{f=1}^{F} \bbA_{d}^f(\bbS) \bbx_{t}^f + \beta_{t} \sum_{d'=1}^{D} \bbB_{d}^{d'}(\bbS) \bbh_{t-1}^{d'} \bigg)\ ,
\end{equation}
where the parameter $\alpha_t \in [0,1]$ is the external input gate, and $\beta_t \in [0,1]$ is the forget gate. We call model \eqref{eqn:GGCRNN} a \emph{Gated Graph Convolutional Recurrent Neural Network} (GGCRNN). Note that $\alpha_t$ adjusts the importance given to the external inputs $\{\bbx_{t}^{f}\}_{f=1}^{F}$ at time $t$ and $\beta_t$ controls how much the GGCRNN \eqref{eqn:GGCRNN} will forget (or, equivalently, remember) from the hidden states $\{\bbh_{t}^{d}\}_{d=1}^{D}$.

Each gate is calculated as the output of a new GCRNN \eqref{eqn:GCRNN} followed by a fully connected layer. To ensure that $\alpha_t$ and $\beta_t$ are well within the unit interval, a sigmoid activation function follows the fully connected layer computations. More precisely, let $\bbmu_{t}^{u} \in \reals^{N}$ be the $U$ graph signals that describe the state of the input gate (alternatively, let $\tbmu_{t}^{i} \in \reals^{U}$ be the $U$ internal values stored at node $i$ that determine the state of the input gate). These states are updated as
\begin{equation}
    \bbmu_{t}^{u} = \xi \left( \sum_{f=1}^{F} \bbGamma^{uf}(\bbS) \bbx_{t}^{f} + \sum_{u=1}^{U} \bbDelta^{uu'}(\bbS) \bbmu_{t-1}^{u'} \right)\ ,
\end{equation}
where $\xi$ is a nonlinear function, and $\bbGamma^{uf}(\bbS)$ and $\bbDelta^{uu'}(\bbS)$ are LSI-GFs, cf. \eqref{eqn:LSIGFs}. The value $\alpha_{t} \in [0,1]$ of the input gate is then calculated by projecting the states $\{\bbmu_{t}^{u}\}_{u=1}^{U}$ of the input gate onto a learned vector $\bbomega \in \reals^{NU}$ and applying a sigmoid
\begin{equation}
    \alpha_{t} = \mathrm{sigmoid}(\bbomega^{\Tr} [(\bbmu_{t}^{1})^{\Tr},\ldots,(\bbmu_{t}^{U})^{\Tr}]^{\Tr})\ .
\end{equation}

Analogously, let $\bbnu_{t}^{v} \in \reals^{N}$ be $V$ graph signals that make up the state of the forget gate. These states are updated as
\begin{equation}
    \bbnu_{t}^{v} = \eta \left( \sum_{f=1}^{F} \bbPhi^{vf}(\bbS) \bbx_{t}^{f} + \sum_{v'=1}^{V} \bbPsi^{vv'}(\bbS) \bbnu_{t-1}^{v'} \right).
\end{equation}
with $\eta$ a nonlinear function, and $\bbPhi^{vf}(\bbS)$ and $\bbPsi^{vv'}(\bbS)$ collections of LSI-GFs, cf. \eqref{eqn:LSIGFs}. Then, the forget gate $\beta_{t} \in [0,1]$ is computed as
\begin{equation}
    \beta_{t} = \mathrm{sigmoid}(\bbtau^{\Tr} [(\bbnu_{t}^{1})^{\Tr},\ldots,(\bbnu_{t}^{V})^{\Tr}]^{\Tr})
\end{equation}
for some learned $\bbtau \in \reals^{NV}$.

Notice that $\alpha_t$ and $\beta_t$ are the same for every node and every feature, but vary with time. This allows exploiting local operations through the graph filters $\bbGamma$, $\bbDelta$, $\bbPhi$ and $\bbPsi$, and keeps the number of learnable parameters under control.


\section{Numerical Experiments} \label{sec:sims}




In this section, we present numerical results obtained using multiple variations of our GCRNN architecture in a synthetic experiment --- ten-step prediction --- and a classification problem involving real seismic data. All simulated architectures consist of a 1-layer GCRNN \eqref{eqn:GCRNN} or GGCRNN \eqref{eqn:GGCRNN} and an output neural network mapping the state $\bbH$ to the target representation $\bbY$, which is either a GNN or a \textit{localized} multi-layer perceptron that mixes each node's local features individually, cf. \eqref{eqn:local_mlp}. In all graph filters, the GSO is the adjacency matrix. The activation function in the GCRNNs is always the hyperbolic tangent, and in the intermediate layers of the output neural network it is the ReLU. In all experiments, the LSI-GFs \eqref{eqn:LSIGFs} have $K = 4$ filter taps. If a GCRNN is gated, the GCRNNs used to compute its input and forget gates have state variables with $U=V=D$ features and $K$ filter taps as well, and are always followed by output neural networks that are full MLPs with a total of $ND$ parameters. All architectures were optimized using ADAM \cite{kingma17-adam} with decaying factors $\beta_{1} = 0.9$ and $\beta_2=0.999$.

\myparagraph{Ten-step prediction} In this experiment, we consider a stochastic block model (SBM) graph $\ccalG$ with $N = 20$ nodes, 4 communities and intra and inter-community edge probabilities of $0.8$ and $0.2$, respectively. Let $\bbS$ be the GSO of $\ccalG$. Given an initial graph signal $\bbx_0 \in \reals^N$, $0 \leq [\bbx_0]_i \leq 1$, the diffused signals $\bbx_t, t = 1,2,\ldots$, are generated as
\begin{equation}
\bbx_t = \bbS \bbx_{t-1} + \bbw_t\ ,
\end{equation}
where $\bbw_t$ is a zero mean gaussian noise that can be correlated both in time and in between nodes. In particular, we chose $\sigma_{\text{time}}^{2} = 0.01$ for the variance across time and $\sigma_{\text{nodes}}^{2} = 0.01$ for the variance across nodes. Likewise, we set the crosscorrelation factors across time and nodes to $\rho_{\text{time}} = \rho_{\text{nodes}} = 0.1$. Fixing the input sequence length to $T = 10$, the $10$-step prediction problem consists of estimating $\bbx_{10}, \bbx_{11}, \dots, \bbx_{19}$ from $\bbx_0, \bbx_1, \dots, \bbx_9$.

We consider 4 GCRNN architectures. The first two are a GCRNN and a Gated GCRNN with $D = 10$ state features whose output neural network is a 1-layer GNN with $K = 4$ filter taps and without fully connected layers. The total number of parameters in these architectures are $480$ and $1,760$ ($480$ for the main GCRNN architecture, and $640$ for those of each gate) respectively. The baseline for comparison is a GNN with 2 convolutional layers and no fully connected layer containing $480$ parameters, the same as the non-gated GCRNN. 

The other two architectures are a GCRNN and a Gated GCRNN where the output neural network is a localized MLP, consisting of the same $D$-parameter MLP per node. These were compared with the Diffusion Convolutional Recurrent Neural Network (DCRNN) from \cite{li2017diffusion} with $1$ recurrent layer, $10$ recurrent units and diffusion length $4$. The number of parameters of the simple GCRNN, the Gated GCRNN and the DCRNN were $450$, $1,730$ and $3,370$ respectively.

All of these architectures were used to simulate the 10-step prediction problem in $10$ rounds, with $10$ different graphs and dataset realizations. In all rounds, the training, validation and test sets comprised $10,000$, $2,400$ and $200$ samples, respectively. All models were trained to optimize the L1 loss (mean absolute error) in 5 training epochs, with batch size 100 and learning rate $0.001$ for the GCRNNs followed by GNNs and $0.005$ for those followed by localized perceptrons. The average test losses and corresponding standard deviations for each architecture are shown in Figure \ref{fig:synthetic}. The main takeaway from Figure \ref{GNN} is that the GCRNN, even when not gated and containing the same number of parameters as the GNN, achieves a considerably smaller loss than the non-recurrent GNN, which attests to the importance of the recurrence mechanism in processing time sequences.
In Figure \ref{MLP}, although the difference among the average test losses achieved by each architecture is minimal, both GCRNN architectures have about a half or less of the $3,370$ parameters of the DCRNN (with the simple GCRNN counting as little as $450$ parameters), and present a much smaller variance as well.

\myparagraph{Earthquake epicenter placement on a seismograph network} The second numerical experiment is a classification problem based on data from GeoNet \cite{geonet}, a geological hazard database from New Zealand, and the Incorporated Research Institutions for Seismology (IRIS) database \cite{iris}, which contains data for 8 active seismographs in the country in the timeframe of analysis. By gathering the origin times of all earthquakes registered by GeoNet between 12/25/2018 and 02/25/2019, we obtain seismic wave readings at these seismographs $30\text{s}$ and $60\text{s}$ before each earthquake, sampled at $2\text{Hz}$. We then construct a 3 nearest-neighbor seismograph network from the seismographs' coordinates, and from the earthquakes' epicenter coordinates we generate labels corresponding to the nearest sensor to the earthquake's epicenter. In this setting, the objective is to accurately predict the closest node to the epicenter of an earthquake from seismic waves collected at the network's nodes immediately before the shock.

The experiment is conducted twice, once for the $30\text{s}$ and once for $60\text{s}$ duration wave. This results in input sequences with length $T=60$ and $T=120$ respectively.
We consider 2 GCRNNs. They are: (i) a GCRNN followed by a GNN with 1 convolutional layer mapping the state features to a single feature, and (ii) its gated version. The number of state features are $60$ and $120$ for the $30\text{s}$ and $60\text{s}$ experiments respectively. In the GCRNNs, the input sequences are used to process a state variable of same length, but only the last state is fed into the output GNN for label prediction. 
The baseline is a GNN with 1 convolutional layer. Because GNNs cannot process sequences, each element of the sequence is interpreted as an input feature. The convolutional layer of this GNN maps $T$ input features to $T+2$ features, which comes down to $4T(T+2)$ parameters because the number of filter taps is always 4. This GNN was chosen to make for a reasonable comparison with the non-gated GCRNN, which has $4T^2 + 8T$ parameters. 

Out of the $2,503$ earthquakes that happened in the two months to which we restrict our analysis, around $80\%$ were used for training and $20\%$ for testing each model. We optimize a cross-entropy loss with learning rate $0.001$ over 10 training epochs and in batches of 100. Test accuracy for each model and each experiment are reported in Table \ref{tb:earthquakes}, as well as the number of parameters in the convolutional layers of each architecture.

\begin{table}[t]
\centering
\begin{tabular}{l|cc|cc} \hline
		& \multicolumn{2}{c}{30-second wave} & \multicolumn{2}{|c}{60-second wave}  \\
Architecture    & Accuracy (\%)  & Param. & Accuracy (\%)  & Param. \\ \hline
GCRNN				& $33.33$ & $14,880$ & $32.93$  & $58,560$ \\ 
Gated GCRNN		& $38.32$ & $29,520$ & $39.12$  & $116,640$ \\
GNN						& $28.34$ & $14,880$ & $30.74$  & $58,560$ \\ \hline
\end{tabular}
\caption{Test accuracy and \# of parameters for each model in the epicenter placement problem for 30s and 60s waves.}
\label{tb:earthquakes}
\end{table} 

Even though all models achieve higher accuracy than random placement, the GCRNN architectures outperform a GNN with same number of parameters as the non-gated GCRNN. Table \ref{tb:earthquakes} also illustrates the importance of the gating mechanism as input sequences grow longer: the percentage difference in the test accuracy achieved by the gated GCRNN and the non-gated GCRNN is 23.8\% bigger in the case of the 60-second wave.


\section{Conclusions} \label{sec:conclusions}



We introduced Graph Convolutional Recurrent Neural Networks (GCRNNs) as NN architectures specifically tailored to deal with problems involving graph processes. Their primary feature is the use of banks of graph convolutional filters to implement the recurrence relationship. Thus, the number of parameters is independent of time and of the size of the graph.
We have further extended this architecture to Gated GCRNNs (GGCRNNs) with input and forget gates that are akin to those of LSTMs.
Numerical results obtained in a synthetic regression problem show that GCRNNs largely improve performance with respect to GNNs when the graph signals are part of a graph process. As for Gated GCRNNs, their ability to take long term dependencies into account was demonstrated in a real world experiment with different input sequence lengths.

\bibliographystyle{IEEEbib}
\bibliography{myIEEEabrv,biblioEusipcoGRNN}

\end{document}